\documentclass{article}
\usepackage[a4paper, portrait, margin=1.0in]{geometry}
\usepackage[english]{babel}
\usepackage[utf8]{inputenc}
\usepackage[T1]{fontenc}
\usepackage{newtxmath}
\usepackage{amsmath}
\usepackage{helvet}
\usepackage{etoolbox}
\usepackage{graphicx}
\usepackage{titlesec}
\usepackage{caption}
\usepackage{xcolor} 
\usepackage{booktabs}
\usepackage{wasysym} 
\usepackage{fdsymbol}
\usepackage{multirow}
\usepackage{hyperref}
\hypersetup{colorlinks, citecolor=cyan}
\usepackage{caption}
\graphicspath{ {./images/} }
\usepackage{fancyhdr}
\usepackage{graphicx}
\usepackage{float}

\fancypagestyle{plain}{
	\fancyhf{}

}
\makeatletter
\patchcmd{\@maketitle}{\LARGE \@title}{\fontsize{16}{19.2}\selectfont\@title}{}{}
\makeatother

\usepackage{authblk}

\newsavebox\affbox
\author[1]{\textbf{Jesudara Omidokun}}
\author[2]{\textbf{Darlington Egeonu}}
\author[3]{\textbf{Bochen Jia}}
\author[4*]{\textbf{Liang Yang}}
\affil[1,2,3]{  Industrial and Manufacturing Systems Engineering Department, University of Michigan, Dearborn, MI, USA.
}
\affil[4]{School of Mechanical and Automotive Engineering, South China University of Technology, Guangzhou 510640, China.
}
\titlespacing\section{0pt}{12pt plus 4pt minus 2pt}{0pt plus 2pt minus 2pt}
\titlespacing\subsection{12pt}{12pt plus 4pt minus 2pt}{0pt plus 2pt minus 2pt}
\titlespacing\subsubsection{12pt}{12pt plus 4pt minus 2pt}{0pt plus 2pt minus 2pt}

\titleformat{\section}{\normalfont\fontsize{10}{15}\bfseries}{\thesection.}{1em}{}
\titleformat{\subsection}{\normalfont\fontsize{10}{15}\bfseries}{\thesubsection.}{1em}{}
\titleformat{\subsubsection}{\normalfont\fontsize{10}{15}\bfseries}{\thesubsubsection.}{1em}{}

\titleformat{\author}{\normalfont\fontsize{10}{15}\bfseries}{\thesection}{1em}{}

\title{\textbf{\huge Leveraging Digital Perceptual Technologies for Remote Perception and Analysis of Human Biomechanical Processes: A Contactless Approach for Workload and Joint Force Assessment}\\
	
 }
\date{}    

\begin{document}

\pagestyle{headings}
\newpage
\setcounter{page}{1}
\renewcommand{\thepage}{\arabic{page}}

\captionsetup[figure]{labelfont={bf},labelformat={default},labelsep=period,name={Figure }}	\captionsetup[table]{labelfont={bf},labelformat={default},labelsep=period,name={Table }}
\setlength{\parskip}{0.5em}

\maketitle

\noindent\rule{16cm}{0.3pt}
\section*{Abstract}
This study presents an innovative computer vision framework designed to analyze human movements in industrial settings, aiming to enhance biomechanical analysis by integrating seamlessly with existing software. 
Through a combination of advanced imaging and modeling techniques, the framework allows for comprehensive scrutiny of human motion, providing valuable insights into kinematic patterns and kinetic data. Utilizing Convolutional Neural Networks (CNNs), Direct Linear Transform (DLT), and Long Short-Term Memory (LSTM) networks, the methodology accurately detects key body points, reconstructs 3D landmarks, and generates detailed 3D body meshes. Extensive evaluations across various movements validate the framework's effectiveness, demonstrating comparable results to traditional marker-based models with minor differences in joint angle estimations and precise estimations of weight and height. Statistical analyses consistently support the framework's reliability, with joint angle estimations showing less than a 5-degree difference for hip flexion, elbow flexion, and knee angle methods. Additionally, weight estimation exhibits an average error of less than 6\% for weight and less than 2\% for height when compared to ground-truth values from 10 subjects. The integration of the Biomech-57 landmark skeleton template further enhances the robustness and reinforces the framework's credibility. This framework shows significant promise for meticulous biomechanical analysis in industrial contexts, eliminating the need for cumbersome markers and extending its utility to diverse research domains, including the study of specific exoskeleton devices' impact on facilitating the prompt return of injured workers to their tasks.
\\
\textbf{\textit{Keywords}}: \textit{motion capture; kinematics; kinetic; computer vision; opensim deep learning; anthropometry; keypoints}
\\
\noindent\rule{16cm}{0.3pt}

\section{Introduction}
Biomechanical analysis is a crucial aspect of workload estimation for healthy workers and rehabilitation for injured and recovering individuals. It provides quantitative and qualitative data to understand the mechanism of human body kinetics and kinematics, allowing for accurate assessment and evaluation of musculoskeletal exposure during various activities\cite{assehr082}. The physical workload imposed by demanding tasks can be rigorously quantified through biomechanical analysis, which utilizes measured kinematic and kinetic parameters, such as joint angles, forces, and torques, to provide objective metrics of the musculoskeletal load experienced by the body\cite{GALLAGHER1994}. Typically, conventional biomechanical analysis methods use motion capture systems and various sensors to gather data on posture, motion, and external forces\cite{Von13131, Shabani2017,Jia2023}. To gather motion data, this procedure usually entails attaching reflective markers to the subject and the markers' positions are captured using the infrared camera system. Simultaneously, additional anthropometric measurements and external forces are usually collected prior to the formal data collection through force gauges and direct body measurements, which furnish the required information for further biomechanical analysis. Subsequently, biomechanical models, encompassing musculoskeletal components such as bones, joints, muscles, ligaments, and cartilage, are employed to compute internal joint loads and muscle forces, etc\cite{Viceconti203}. Despite their utility, conventional biomechanical analysis methods face several limitations. As noted by Von Marcard et al.\cite{Von13131} and Mehrizi et al.\cite{Shabani2017}, these techniques typically require expensive equipment and controlled laboratory environments, making them impractical for real-world workplace assessments. Additionally, the complex and invasive nature of these techniques, such as marker placement and its irritability to the skin, can pose challenges for participants and limit their widespread adoption in actual work settings.

The rapid advancement of AI technology offers possible solutions for problems associated with conventional biomechanical analysis techniques. One such avenue is computer vision (CV), a multidisciplinary field that empowers machines to "see" and understand the visual world through computational models and cameras\cite{Yu2019, Yu2019a, Shabani2017, Wei96}. This transformative technology has garnered significant traction in biomechanics, particularly in the realm of pose estimation for human kinematic analysis.

CV algorithms have been instrumental in developing innovative approaches for extracting 2D or 3D joint coordinates from recordings of one or more individuals using cameras\cite{Schmidhuber2015, LeCun2015, Huang1996, Fang2020}. These algorithms analyze visual data to identify key body landmarks, such as joints and extremities, and estimate their spatial positions with remarkable accuracy. This advancement eliminates the need for cumbersome reflective markers often used in traditional motion capture systems, significantly enhancing flexibility and ease of use.

Furthermore, CV-based approaches offer several advantages over conventional techniques: \begin{enumerate}
    \item[\textbullet] Accessibility\cite{trucco1998introductory}: Camera-based systems are considerably less expensive and more readily available compared to specialized motion capture equipment, democratizing access to biomechanical analysis for researchers and clinicians alike.
    \item[\textbullet] Real-world applicability\cite{forsyth2012computer}: Unlike the controlled laboratory environments required for traditional methods, CV can be readily deployed in real-world settings, enabling on-site assessments of physical workload and posture in workplaces, rehabilitation centers, and even homes.
    \item[\textbullet] Non-invasive nature\cite{szeliski2010computer}: Markerless pose estimation eliminates the need for attaching sensors to the subject, improving comfort, and reducing potential interference with movement patterns.
    \item[\textbullet] Data richness\cite{prince2012computer}: Beyond mere joint positions, advanced CV algorithms can extract additional information from video recordings, such as limb velocities, accelerations, and even muscle activations, providing a more holistic understanding of human kinematics and kinetics.
\end{enumerate}

\begin{figure}[H]
    \centering
    \includegraphics[width=1\textwidth]{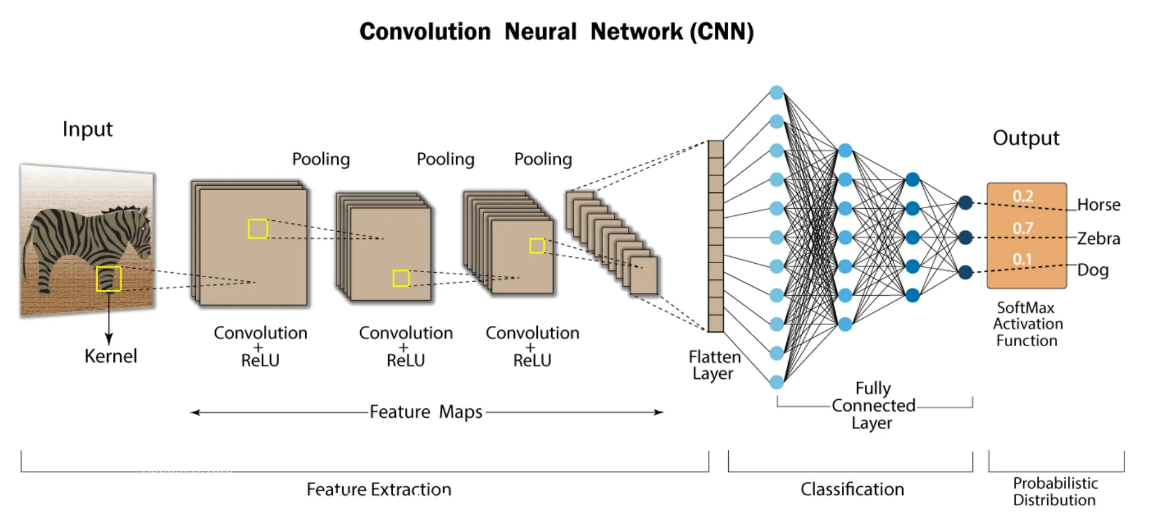}
    \caption{An example of CNN architecture\cite{Demertzis2023}}
    \label{fig:fig1}
\end{figure}

\begin{figure}[H]
    \centering
    \includegraphics[width=1\textwidth]{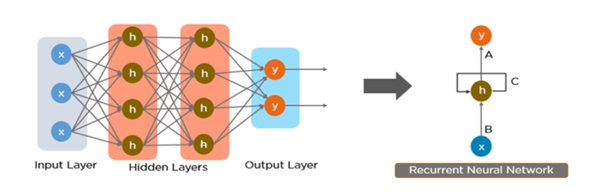}
    \caption{ A typical RNN architecture\cite{Banerjee2019}}
    \label{fig:fig2}
\end{figure}

The field of pose estimation has undergone a significant transformation with the rise of deep learning, specifically through the widespread adoption of convolutional neural networks (CNNs) and recurrent neural networks (RNNs)\cite{Yu2019, Wei96, Guo2016DeepLF}. These advanced algorithms, capable of learning intricate patterns from vast datasets, have revolutionized the extraction and analysis of human body positions from visual information. Guo et al.\cite{Demertzis2023} describe the typical architecture of a CNN, consisting of input, hidden, and output layers, with convolutional and pooling layers in the hidden layers. These layers act as feature detectors, extracting relevant information and spatial relationships within the input image, ultimately translating them into 3D joint coordinates. Figure \ref{fig:fig1} and Figure \ref{fig:fig2}  illustrate a typical RNN and CNN architecture, which includes an input layer of image data, multiple mapping of convolutional and pooling layers within the hidden layers for feature extraction and classification, and an output layer for predicting or estimating the intended result. This could involve categorizing the input image into a specific class, as illustrated in Figure \ref{fig:fig1}, or generating an arbitrary output, as demonstrated in Figure \ref{fig:fig2}.

While CNNs excel in spatial data analysis like images, RNNs are particularly effective in understanding temporal dynamics, making them ideal for processing the sequential nature of human movement. With loops incorporated into their architecture, RNNs can analyze time series data during feature extraction, capturing subtle changes in joint positions over time and leading to more accurate 3D pose estimation\cite{Guo2016DeepLF}. The choice between CNNs and RNNs depends on factors such as the application and available dataset. CNNs are preferred for 2D pose estimation, focusing on joint locations within a single frame, while RNNs shine in 3D pose estimation, understanding joint movement over time. Despite both CNNs and RNNs utilizing complex network architectures, their approaches differ significantly. CNNs are feedforward networks relying on filters and pooling layers, while RNNs employ a feedback mechanism, processing information sequentially and utilizing previous outputs. Additionally, CNNs handle fixed-size inputs and outputs, whereas RNNs can process variable-length data, making them suitable for video sequences with varying frame numbers. In summary, CNNs and RNNs are complementary tools within the deep learning arsenal, each with unique strengths for specific pose estimation challenges. Leveraging their capabilities and understanding their nuances enables researchers to unlock the full potential of these algorithms for more accurate and insightful analyses of human movement.

The convergence of biomechanics and computer vision has ignited a revolution in analysis of human behaviors, opening doors to previously unimaginable avenues for ergonomic analysis and intervention in diverse industry settings, such as construction, healthcare, and even daily activities. A plethora of tools and frameworks are available, each catering to specific needs and offering distinct capabilities. OpenPose\cite{Cao2019}, for instance, shines in real-time multi-person key points detection, while DeepLabCut\cite{Mathis2018} leverages deep learning for user-defined point tracking. Frameworks such as PyCaret\cite{Arai2023} and TensorFlow/PyTorch\cite{Usvyatsov2022} tackles machine learning-driven analysis by offering pre-trained models and customizable algorithms. Additionally, Simi Motion\cite{Markerless3D_2024} is recognition for its versatility in seamlessly integrating video-based analysis with robust tools designed for sports performance evaluation.

Integration with the aforementioned computer vision (CV) tools and frameworks presents an opportunity to leverage established human biomechanical modeling tools, such as OpenSim\cite{Delp2007} and AnyBody\cite{Trinler2019}, to create comprehensive solutions for biomechanical analysis of the musculoskeletal system. These combined CV-based biomechanical models could offer valuable real-time insights into an individual's musculoskeletal state, thereby minimizing the potential risk of injury during various tasks across diverse industry settings, including healthcare, construction, and manufacturing (e.g.,\cite{pavllo_3d_2019,Yu2019, Pagnon2021}). These frameworks empower the development of intelligent algorithms and models that could assess ergonomic risks, quantify force exertion, and even personalize injury prevention strategies.
\subsection{Statement of the Problem}
Conventional biomechanical analysis methods use physical markers or tracking devices, which can be intrusive, time-consuming, and potentially uncomfortable for the individual. Additionally, these methods require specialized equipment and expertise, making them less accessible in specific settings and expensive. Consequently, there is a recognized need for a more accessible and non-intrusive approach to biomechanical analysis capable of delivering precise and comprehensive data on human movement.

While emerging methods employing computer vision techniques show promise in biomechanical analysis, there remains room for improvement in terms of accuracy, real-time processing, and automated analysis. Notably, techniques such as markerless pose estimation and deep learning algorithms hold potential for addressing these challenges and propelling the field of biomechanics forward through the integration of computer vision.

Despite the availability of various computer vision methods like OpenPose\cite{Cao2019}, DeepLabCut\cite{Mathis2018}, VideoPose3D\cite{pavllo_3d_2019}, and Pose2Sim\cite{Pagnon2021}, a critical gap exists in providing an integrated system with seamless compatibility and automated analysis alongside established biomechanical analysis tools and software, such as OpenSim.

While Egeonu and Jia\cite{Egeonu2024} provide a comprehensive overview of various integration levels between computer vision and biomechanical models, existing approaches often necessitate additional steps involving direct and potentially invasive measurements using various sensors\cite{Pagnon2021,Wang2021,mehrizi_predicting_2019,Wei96,wang_load_2021}. Researchers often face extra tasks of manually processing and analyzing data obtained from these steps. This highlights the need for further research on seamless integration methods that eliminate the reliance on such extraneous measurements, thereby streamlining the overall analysis process.

Therefore, the aim of this study is to establish an advanced computer vision framework tailored specifically for advanced biomechanical analysis, seeking to obviate the necessity for direct and potentially invasive measurements while ensuring the provision of essential data requisite for the comprehensive analysis of human body dynamics, which includes:
\begin{enumerate}
    \item [\Circle] Eliminating the need for direct and invasive techniques: By leveraging state-of-the-art computer vision algorithms, the framework extracts crucial data for biomechanical analysis without the need for physical markers, sensors, or direct force measurements. This not only enhances accessibility and comfort for participants but also expands the applicability of the framework to real-world settings beyond controlled laboratory environments.
    \item [\Circle] Providing comprehensive biomechanical data: The framework go beyond tracking movement. By analyzing visual cues and incorporating advanced algorithms, it derives a comprehensive set of biomechanical parameters, including:
          \begin{enumerate}
              \item [\(\smallblacksquare\)] Anthropometric data: Limb lengths, segmental masses, body composition, and other relevant anthropometric measures will be seamlessly acquired through visual data, eliminating the need for additional measurement tools.
              \item [\(\smallblacksquare\)] Kinematic data: Joint angles, velocities, accelerations, and range of motion will be precisely tracked, providing detailed insights into joint function and movement patterns.
              \item [\(\smallblacksquare\)] Kinetic data: Gravitational external forces based on body weight, facilitates the estimation of segment forces. This analytical approach offers valuable insights into external loads and forces.
          \end{enumerate}

\end{enumerate}

\section{Methodology}

\begin{figure}[H]
    \centering
    \includegraphics[width=1.\textwidth]{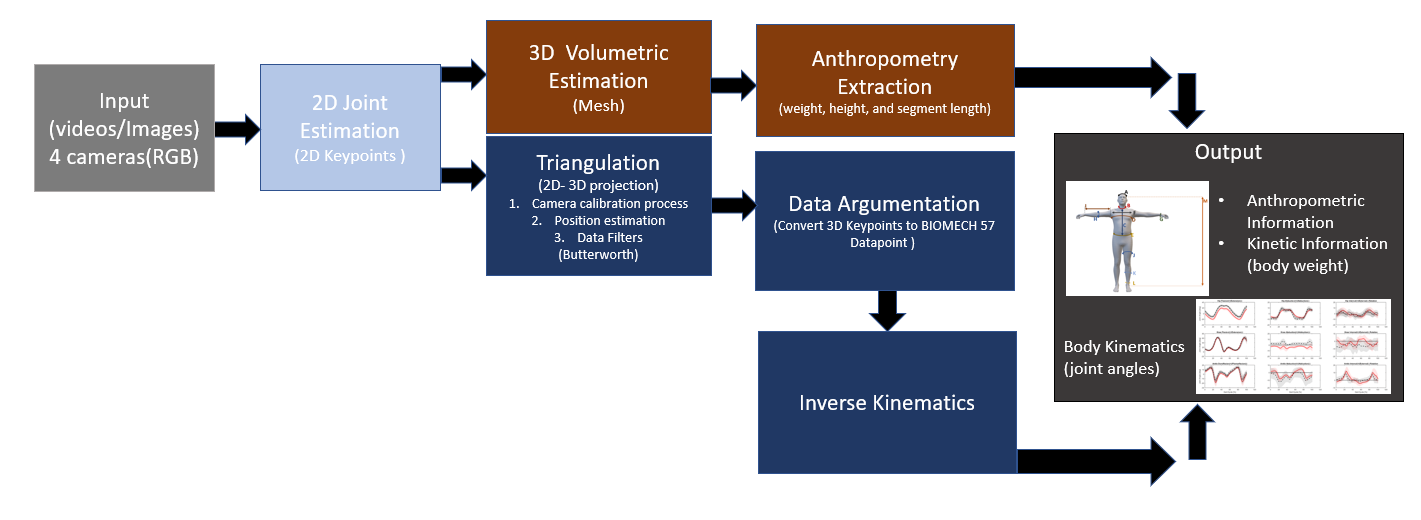}
    \caption{The structure of the developed framework}
    \label{fig:fig3}
\end{figure}

The computer vision framework for biomechanical analysis comprises essential components and algorithms that collaborate to achieve precise tracking analysis of movements without the reliance on physical markers and anthropometric measurements. Through the synergistic integration of proven methodologies, including OpenPose and SMPL-X, this study introduces an innovative computer vision framework designed to overcome existing limitations, e.g., insufficient number of landmarks, and provide comprehensive and robust results.

This section delineates the methodology flow illustrated in Figure \ref{fig:fig3}, elucidating the step-by-step process initiated with the acquisition of video data, serving as input for a multi-stage CNN architecture dedicated to 2D joint estimation and detection. The 2D joint estimation process involves identifying key points on the human body within a 2D image, corresponding to joints such as elbows, wrists, shoulders, hips, knees, and ankles. Optical flow and Kalman filtering techniques are employed to extract and track joint estimation information from video frames. The derived 2D joint estimation data serves as input for subsequent 3D volumetric estimation and 3D projection processes. In the 3D volumetric estimation process, the estimated 2D joint positions are utilized to reconstruct a 3D mesh model of the human body. This is achieved through a CNN architecture that combines various blend shape and pose shape functions, providing a three-dimensional representation facilitating accurate measurement of body segment volume and shape. The 3D mesh model, integrated with a weight/height estimation algorithm, offers insights into body composition and anthropometric measurements. The 3D projection process maps the 3D mesh model onto the camera frame coordinate system, enabling visualization of the human body. The resulting 3D pose data are further aligned with the OptiTrack skeleton marker set template (Biomech-57)\cite{optitrack23} using a trained Long Short-Term Memory (LSTM\cite{Lindemann2021}) neural network model. Integration of this computer vision framework with biomechanical analysis software, such as OpenSim, ensures seamless data integration and analysis. The system block diagram outlines the comprehensive process employed to achieve the primary objective of extracting anthropometric details, including weight, height, segment information, marker trajectories, and ground reaction forces. Moreover, the framework incorporates modules for joint angle calculation, range of motion analysis, and movement pattern assessment, offering a holistic approach to biomechanical analysis.
\subsection{Input}
The computer vision methodology for biomechanical analysis is tailored to scrutinize and interpret visual inputs derived from video or image data, extracting essential information pertaining to human kinematic and kinetic parameters. The requisite input for the computer vision framework encompasses video data capturing human body data and movements, coupled with pertinent camera calibration information. The video data may encompass recordings of a subject executing specific movements or a sequence of images capturing motion frames. Crucially, the video must offer clear and consistent imaging, providing a nearly 360° view of the subject’s body to ensure precise tracking and analysis of the subject’s movements. The system integrates four strategically positioned cameras to capture a comprehensive view of the subject. This multi-camera setup facilitates optimal coverage of the subject’s body, minimizing occlusions that could impede accurate tracking and analysis. The angles at which the cameras are installed are carefully chosen to enhance the overall visibility of the subject. This setup ensures a robust and reliable data capture process, essential for the subsequent biomechanical analysis.

\subsection{2D Joint Estimation}
\begin{figure}[H]
    \centering
    \includegraphics[width=0.4\textwidth]{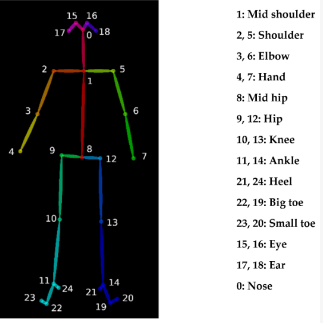}
    \caption{The landmarks (keypoints) representation Body\_25B model used in current study\cite{cao2017realtime}.}
    \label{fig:fig4}
\end{figure}

Following the acquisition of input data, the computer vision framework undertakes the processing of video data to extract crucial information, specifically 2D joint estimations. This framework eliminates the need for physical markers, leveraging deep learning methods to directly derive 2D joint information from the video data. Human skeletal keypoints, or landmarks, are identified from image frames (refer to Figure \ref{fig:fig4}) utilizing a CNN model, adapted and refined from an existing approach.

The model architecture operates through predicting the heatmap of landmarks through a multi-stage network. Subsequently, various techniques and algorithms are employed to refine the predictions, including part affinity fields (PAFs), Non-Maximum Suppression and Greedy Inference, and Bipartite Matching. PAFs utilizes CNNs to process input images in multiple stages, making predictions and refining those predictions iteratively. Convolutional, pooling, and fully connected layers are applied to an image to produce part affinity fields (PAFs) and confidence maps.

\begin{figure}[H]
    \centering
    \includegraphics[width=1.\textwidth]{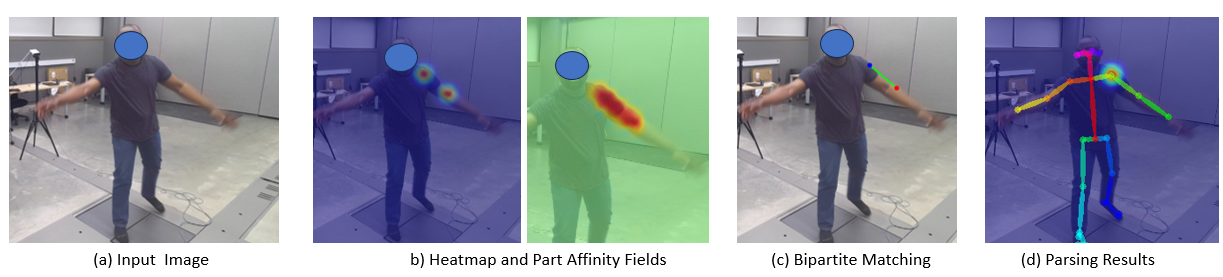}
    \caption{A schematic of part affinity fields (PAFs) with the input image to the parsing result.}
    \label{fig:fig5}
\end{figure}

The confidence map serves as a probability density function on the new image, indicating the likelihood that the pixel color will occur in the object in the previous image. PAFs provide geometric information about the pose direction and location of limbs, effectively establishing connections between body parts, as indicated in Figure \ref{fig:fig5}. These connections act as vectors for each pixel, connecting body parts like shoulders to elbows, ensuring accurate correspondence of joint locations\cite{Cao2019}. Subsequently, the system employs Non-Maximum Suppression to identify local maximum points, signifying potential body part locations, which are designated as 'candidates.' A greedy algorithm is then applied to establish connections between candidates, approximating a minimum spanning tree. Bipartite matching is employed to confirm candidate parts, which are then organized into a set of bipartite matches, culminating in the final pose estimation.

The framework introduced in this study extends the capabilities of 2D poses estimations by enhancing its functionality to accurately estimate 3D poses through the utilization of the Direct Linear Transform (DLT) method. Additionally, this framework integrates supplementary algorithms dedicated to tracking analysis and filtering, aligning with the specific objectives of the study.

The framework's contribution is further emphasized by its amalgamation of 2D pose estimation with auxiliary framework to addresses the deficiencies in mesh estimation inherent in the 3D shape estimation process. The primary goal is to attain precise anthropometric measurements in alignment with the study's objectives. Detailed discussions elaborating on the solutions devised to overcome these limitations are presented in subsequent sections.

\subsection{The Anthropometry Information}
\begin{figure}[H]
    \centering
    \includegraphics[width=1.0\textwidth]{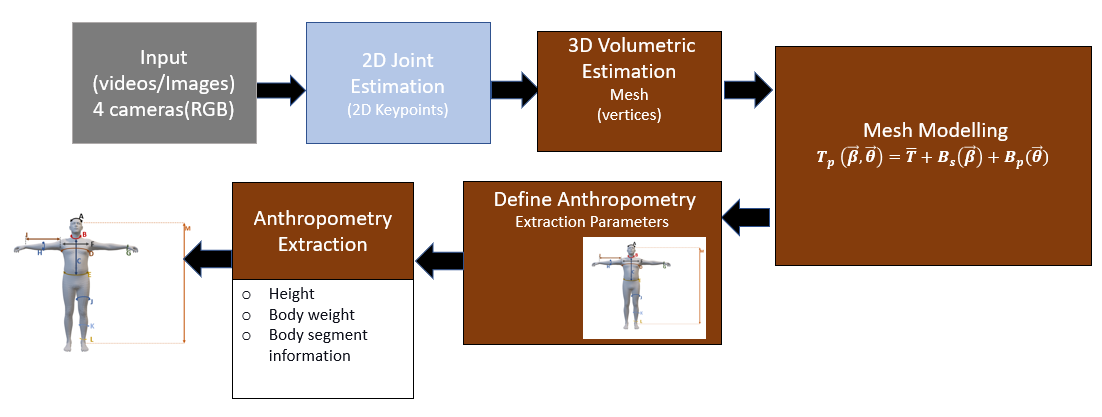}
    \caption{Block Diagram Illustrating Anthropometry Extraction Algorithms}
    \label{fig:fig6}
\end{figure}

Anthropometry information consists of estimating the height, weight, and body segment information from each subject. The block diagram in Figure \ref{fig:fig6} shows the process used to achieve the required objective of extracting weight, height, and segment information.
\subsubsection{3D Volumetric Estimation} \label{volume3d}
The methodology involves adapting a template mesh based on the SMPL-X model, introducing offsets to represent new body shapes and pose-dependent changes. The mesh vertices play a pivotal role in determining joint positions for a given body shape. This mesh template incorporates expressive hands and faces, alongside the body shape, utilizing a deep learning approach. It is a generative 3D model that amalgamates body shape, pose, and blend shapes, portraying a diverse range of human body forms. The algorithm employs blend shapes and base mesh deformations, combining them to create a variety of body forms. Two essential blend shape functions, blend shapes (BS) and pose blend shapes (PBS), are integrated into the model. Blend shapes capture identity-related variations in body shape across multiple individuals, while pose blend shapes depict how body shape changes with different poses.

The study outlines the landmark position ($\theta$) with a skinned body model defining vertices of a template ($\bar{T}$) in rest pose joint positions ($J$) and blend weights ($W$) using blend shapes. The algorithm automatically calculates the influence of each blend shape based on the body's pose, thereby correcting pose-dependent shape variation, and addressing standard skinning errors. Additionally, the model algorithm applies principle component analysis (PCA) to handle complex pose-dependent deformations in blend shapes. Furthermore, a Jaw Kinematic model is incorporated for accurate representation of jaw movement, contributing to the enhanced realism of human facial modeling.

\begin{figure}[H]
    \centering
    \includegraphics[width=1.\textwidth]{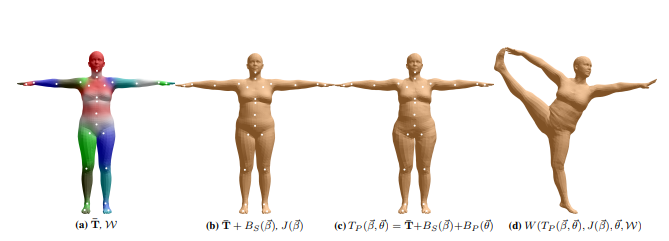}
    \caption{(a) Template mesh with colored blend weights and white-colored joints. (b) Identity-driven blend shape contribution; the shape vector $\vec{\beta}$ vertex and joint locations are linear. (c) Notice how the hips have expanded with the addition of pose blend shapes to prepare for the split pose. (d) Using dual quaternion skinning split pose deformed vertices reposed\cite{Bogo2016}.}
    \label{fig:fig7}
\end{figure}

The mesh is derived from a template (or mean) mesh, denoted as $\bar{T}$, extracted from a database comprising over 1,700 3D scans of 44 distinct subjects in various poses. Subsequently,  $\bar{T}$ is combined with the offset of vertices corresponding to the body shape and pose blend, represented by  $\vec{\beta}$  and $\vec{\theta }$ vectors, respectively. The offsets for both body shape and pose blend are expressed as $B_s\left(\vec{\beta}\right)$ and $B_p \left(\vec{\theta }\right)$, respectively. The outcome of this addition signifies the emergence of new body shapes and pose-dependent shape changes, denoted as $T_p\ \left(\vec{\beta },\vec{\theta }\right)$, referred to as PBS. The pipeline illustrating the progression from the template to the PBS is depicted in Figure \ref{fig:fig7}.

The final stage in the process involves transitioning from PBS to the skinning stages, accounting for joint deformation $J\left(\vec{\beta }\right)$. At this stage, a weighted combination ($W$) of joint deformation $J\left(\vec{ \beta }\right)$ is applied to each mesh vertex from the Pose Blend Shape $T_p\ \left( \vec{\beta },\vec{\theta }\right)$. Consequently, the ultimate stage in mesh formation is skinning $W\left(T_p\left(\vec{\beta },\ \vec{ \theta }\right),\ J\left(\vec{\beta }\right),\ \vec{\theta}\ ,\ W\right)$ , wherein the mesh undergoes joint rotation or vertex alteration more prominently based on the proximity of the vertex.

\subsubsection{Weight and Height Estimation Algorithms}
\begin{figure}[H]
    \centering
    \includegraphics[width=0.7\textwidth]{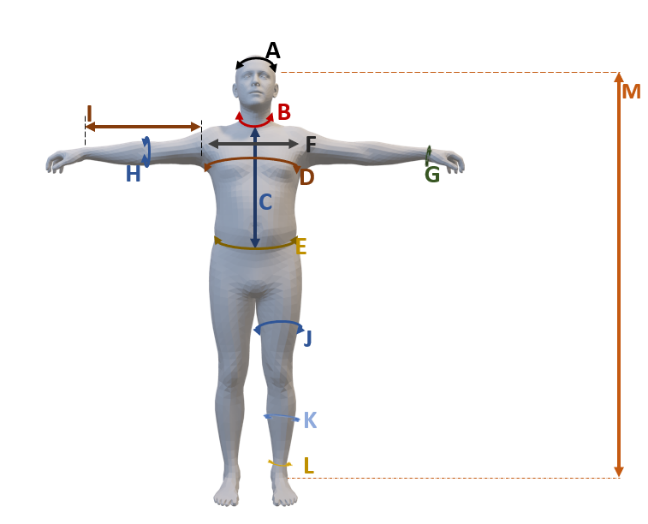}
    \caption{Visual Representation of Anthropometric Extraction Definition}
    \label{fig:fig9}
\end{figure}

The anthropometry extraction algorithm leverages a mesh generated from the section \ref{volume3d} to derive precise measurements of various body dimensions, including height, weight, segment length, and circumference. The index of the mesh vertices identifies and locates anatomical landmarks on the body, as illustrated in Table \ref{tab:table2}, facilitating precise measurement definition parameters shown in Figure \ref{fig:fig9}.
Considering variations in body shape, size, and posture, the algorithm ensures that the extracted anthropometric measurements accurately reflect individual biomechanical characteristics. Operating based on the provided vertices (N*3, where N is the number of vertices and 3 indicates Cartesian coordinates) from individual segment parts, the algorithm utilizes a mesh representation comprising vertices, edges, and faces. This compilation collectively defines the shape and surface of a three-dimensional object in computer graphics and computational geometry. Vertices denote points in 3D space, edges connect pairs of vertices, and faces are polygons formed by connecting edges in a specific sequence. The mesh generates N = 10,475 vertices, with 54 joints and body segments defined based on the index corresponding to the part identified with the joint's keypoints.
\begin{table}[H]
    \centering
    \caption{The measurements name}
    \label{tab:table1}
    \begin{tabular}{|c|c|} \hline
        \textbf{} & \textbf{Measurement Name}   \\ \hline
        A         & Head Circumference          \\ \hline
        B         & Neck Circumference          \\ \hline
        C         & Full Torso Length           \\ \hline
        D         & Chest Circumference         \\ \hline
        E         & Waist Circumference         \\ \hline
        F         & Shoulder to Shoulder length \\ \hline
        G         & Wrist Circumference         \\ \hline
        H         & Forearm Circumference       \\ \hline
        I         & Arm Length                  \\ \hline
        J         & Thigh Circumference         \\ \hline
        K         & Calf Circumference          \\ \hline
        L         & Ankle Circumference         \\ \hline
        M         & Height                      \\ \hline
        N         & Weight                      \\ \hline
        O         & Torso Weight                \\ \hline
        P         & Arm Left Weight             \\ \hline
        Q         & Arm Right Weight            \\ \hline
    \end{tabular}
\end{table}

Utilizing the available mesh information, a total of 17 body measurements (A - Q) are defined, as detailed in Table \ref{tab:table1}. Measurements A - L are standard and have been utilized in prior studies\cite{Pujades2019,Yan2021,yan2021learning}, while L - Q are specifically related to the system integration within the presented framework. The calculation of height and length is based on corresponding vertices and landmarks, such as determining full body height from the head top landmark to the left heel landmark. The Euclidean distance formula is employed to calculate the length between landmark indices.
\begin{eqnarray}
    \mathrm{Euliden\ distance}\left(\mathrm{l}\right)\mathrm{=}\sqrt{\mathrm{(}{\mathrm{x}}_{
            \mathrm{2}}\mathrm{-}{\mathrm{x}}_{\mathrm{1}}{\mathrm{)}}^{\mathrm{2}}\mathrm{+}{
            \mathrm{(}{\mathrm{y}}_{\mathrm{2}}\mathrm{-}{\mathrm{y}}_{\mathrm{1}}{\mathrm{)}}^{
                \mathrm{2}}\mathrm{+(}{\mathrm{z}}_{\mathrm{2}}\mathrm{-}{\mathrm{z}}_{\mathrm{1}}
            \mathrm{)}}^{\mathrm{2}}}
    \label{Eq1}
\end{eqnarray}

Where $\left({\mathrm{x}}_{\mathrm{1}},{\mathrm{y}}_{\mathrm{1}},{\mathrm{z}}_{\mathrm{1}} \right)$ is the coordinate point of the first landmarks and $\left({\mathrm{x}}_{ \mathrm{2}},{\mathrm{y}}_{\mathrm{2}},{\mathrm{z}}_{\mathrm{2}}\right)$ is the coordinate point of the second landmarks.
\begin{table}[H]
    \centering
    \caption{: The landmarks and their corresponding vertex Index}
    \label{tab:table2}
    \begin{tabular}{|p{1.6in}|p{0.7in}|p{0.9in}|} \hline
        \textbf{Landmark Name} & \textbf{Vertex Index} & \textbf{Landmark Index}\\ \hline
        HEAD
        TOP                    & 8976                  & 0                       \\ \hline
        HEAD LEFT TEMPLE       & 1980                  & 16                      \\ \hline
        NECK ADAM APPLE        & 8940                  & 1                       \\ \hline
        LEFT HEEL              & 8846                  & 21                      \\ \hline
        RIGHT HEEL             & 8635                  & 24                      \\ \hline
        SHOULDER TOP           & 5616                  & 2,5                     \\ \hline
        RIGHT WRIST            & 7449                  & 4                       \\ \hline
        LEFT WRIST             & 4823                  & 7                       \\ \hline
        LEFT ELBOW             & 4219                  & 6                       \\ \hline
        RIGHT ELBOW            & 6788                  & 3                       \\ \hline
        LEFT SHOULDER          & 4442                  & 5                       \\ \hline
        RIGHT SHOULDER         & 7218                  & 2                       \\ \hline
        RIGHT HIP              & 3732                  & 9                       \\ \hline
        LEFT HIP               & 4112                  & 12                      \\ \hline
        LEFT\_ANKLE            & 5880                  & 11                      \\ \hline
        RIGHT\_ANKLE           & 3732                  & 14                      \\ \hline
    \end{tabular}
\end{table}

To extract circumference, the landmark index corresponding to the measurement region serves as the starting point for slicing the mesh at the vertex points or landmark position, followed by summing up the resulting line segments\cite{yan2021learning,Yan2020}. Weight is estimated from the mesh volume of the entire body or region, employing the concept of a convex polyhedron. The volume is calculated using the Shoelace Formula,
\begin{eqnarray}
    \mathrm{\ Volume\ (}{\mathrm{V}}_{\mathrm{extracted}}\mathrm{\ }\mathrm{)\ }\mathrm{=}
    \frac{\mathrm{1}}{\mathrm{6}}\left|\sum^{\mathrm{n}}_{\mathrm{i=1}}{\left({\mathrm{x}}_{
            \mathrm{i}}{\mathrm{y}}_{\mathrm{i+1}}\mathrm{-}{\mathrm{x}}_{\mathrm{i+1}}{\mathrm{y}}_{
            \mathrm{i}}\right)}\mathrm{\cdot }{\mathrm{z}}_{\mathrm{i}}\right|
    \label{Eq2}
\end{eqnarray}

Here, $\left({\mathrm{x}}_{\mathrm{I}}\mathrm{,\ }{\mathrm{y}}_{\mathrm{I}}\mathrm{, \ }{\mathrm{z}}_{\mathrm{I}}\right)$ are the coordinates of the vertices in the 3D space, and the indices are taken from vertices points. The body weight is estimated based on human body density, considering the World Health Organization (WHO) body mass index (BMI) indicator for variations in body density with respect to weight categories: underweight (BMI less than 18.5), normal weight (BMI 18.5 to 24.9), overweight (BMI 25 to 29.9), and obese (BMI 30 or greater).
\begin{eqnarray}
    {\mathrm{w}}_{\left\{\mathrm{extracted}\right)}\mathrm{=}\mathrm{\rho}\mathrm{
        \cdot }{\mathrm{V}}_{\mathrm{extracted}}
    \label{Eq3}
\end{eqnarray}

Where $w_{extracted}$ is the weight used in the BMI calculation, $\rho $ is value of human body density at 985 $\frac{kg}{m^3}$\cite{Pujades2019,Yan2021}and $V_{extracted}$ is volume extracted from the mesh in m${}^{3}$.
\begin{eqnarray}
    BMI_{cal}=\frac{W_{extracted}\left(kg\right)}{estimated\ height\left(m^2\right)}
    \label{Eq4}
\end{eqnarray}

The difference between the BMI calculated and WHO BMI category median is added or subtract to the weight extracted $w_{extracted}$.
\begin{eqnarray}
    w=w_{extracted}+\left(BMI_{cal}-BMI_{median}\right)h^2_{est}
    \label{Eq5}
\end{eqnarray} \\

Here, the weight $w$ is estimated, either considering the entire body or a specific body segment. $BMI_{median}$ represents the median value of the BMI category to which $BMI_{cal}$ belongs, and $h_{est}$ denotes the estimated length for either the entire body or the segment of the body.

\subsection{Marker Trajectories}\label{3D_traget}
\begin{figure}[H]
    \centering
    \includegraphics[width=1.\textwidth]{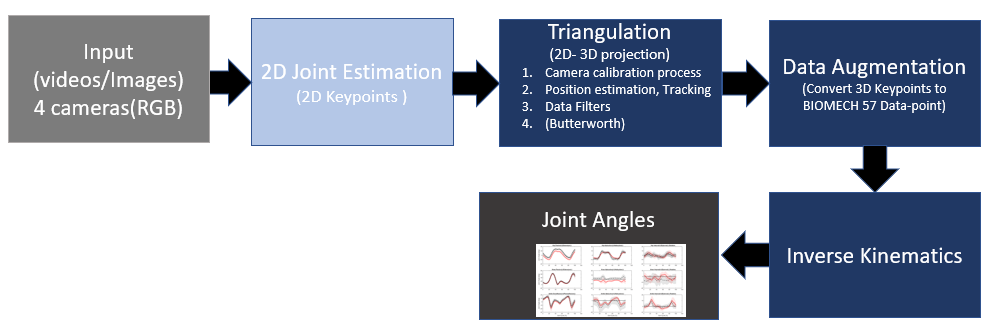}
    \caption{A block diagram of the motion capture system with emphasis on marker trajectory}
    \label{fig:fig10}
\end{figure}

Marker trajectories constitute a pivotal element in biomechanical analysis, furnishing essential information on the movement and positioning of specific anatomical landmarks or markers on the body throughout various activities. The extraction of marker trajectories in the absence of a visible marker on the subject is referred to as marker trajectory. Figure \ref{fig:fig10} illustrated the marker trajectory process, encompassing the transition from input data to the estimation of joint angles.

The marker trajectory process comprises three fundamental steps: first, the application of 3D triangulation utilizing predefined marker points; second, data augmentation to align with a specified template; and third, meticulous filtering or smoothing to mitigate noise and ensure the precision of data localization. This section commences with an exploration of the 3D triangulation process, subsequently delving into an examination of the data augmentation process and culminating in an analysis of the ensuing inverse kinematics process.

\subsubsection{3D triangulation}\label{3D_target1}
The 3D triangulation process involves identifying the x, y, and z coordinates of markers on the body to define their positions relative to the origin of a specified coordinate system. Conventionally, this is achieved by placing reflective markers at specific joints and positions on the subject's body, and camera-based motion capture systems, with a calibration file, to track and record the markers' positions. This establishes the 3D coordinates of the markers in relation to the camera-based motion capture system's defined origin.

In this study, a 3D triangulation method was developed to build the 3D coordinate system in a non-invasive way. It reconstructs synchronized 2D landmarks or key points from 2D joint estimation data using two computer vision algorithms: Direct Linear Transformation (DLT) and singular value decomposition. The DLT triangulation\cite{hartley1997triangulation} method combines the weighted contribution of each key points' confidence score with their corresponding data to estimate the 3D positions. The DLT process determines the rotation matrix (R), the location of the projection center K${}_{0}$, and the calibration matrix (K) from six or more points in the image. To determine R, K${}_{0}$${}_{,}$ and K, the process begins with establishing a relationship between the 2D pixel coordinate and the 3D world coordinate:
\begin{eqnarray}
    \mathrm{\mu }\boldsymbol{\mathrm{k}}\mathrm{=}\boldsymbol{\mathrm{AK}}\boldsymbol{
        \mathrm{\ }}\
    \label{Eq6}
\end{eqnarray}
Where $\boldsymbol{\mathrm{k}}\mathrm{=}\left(\mathrm{X},\mathrm{Y}\mathrm{,1}\right)$ is the coordinates of a 2D object point on camera image also known as 2D pixels, $\boldsymbol{ \mathrm{k}}\mathrm{=}\left(\mathrm{X},\mathrm{Y}\mathrm{,1}\right)$  is identified from the image. While  $\mathrm{\mu }$ ausing ais an unknown scale factor, which is usually represented as one when using the actual pixel scale from the image. Further in equation \ref{Eq4},  $\boldsymbol{\mathrm{K}}\mathrm{=}\left(\mathrm{x},\mathrm{y},\mathrm{z} \mathrm{,1}\right)$ is the coordinates of a 3D object point in 3D world coordinate. While A is the projection matrix which is an unknown parameter and expressed in homogenous coordinate in a 3*4 matrix. For example, the row of  $\mathrm{A=}\left({\mathrm{A}}_{\mathrm{1}}\mathrm{T,}{\mathrm{A}}_{\mathrm{2}}\mathrm{T,}{\mathrm{A}}_{\mathrm{3}}\mathrm{T,}{\mathrm{A}}^{\mathrm{T}}_{\mathrm{4}}\right)$

\begin{eqnarray}
    A=K\left[
    \begin{array}{cc} 
        R & -R K_0 \\
        0^T & 1
    \end{array}\right]
    \label{Eq7}
\end{eqnarray} $A$ allows to determine the rotation matrix $(R)$, the location of the projection center $K_{0}$, and the calibration matrix $(K)$. To solve for $A$, since we know that $A$ is a homogeneous system, therefore:
\begin{eqnarray}
    \ \ AK\ =0\
    \label{Eq8}
\end{eqnarray}

\begin{eqnarray}
    \Longrightarrow \left(A^{\top }_1-XA^{\top }_3\right)\boldsymbol{K}=0,\left(A^{\top
    }_2-YA^{\top }_3\right)\boldsymbol{K}=0
    \label{Eq9}
\end{eqnarray}

Simplify:
\begin{eqnarray}
    \mathrm{\Longrightarrow }{{\mathrm{a}}_{\mathrm{x}}}^{\mathrm{T}}\boldsymbol{\mathrm{K}}
    \mathrm{=0,}{{\mathrm{a}}_{\mathrm{y}}}^{\mathrm{T}}\boldsymbol{\mathrm{K}}
    \label{Eq10}
\end{eqnarray}

For every point in the image, the homogeneous system equation \ref{Eq10} was obtained. The two homogenous equation contains a coefficient vector $\left(\mathrm{a}\right)$ and vector of unknown $\left(\mathrm{K}\right)$ for both x and y dimension. These vectors are then multiplied together, resulting in an equation that equals zero. Using singular value decomposition (SVD) to solve for unknown K, where the 3D object point K has a least square solution. The process for solve for A can be done by writing as $\mathrm{A=X} \mathrm{\theta }{\mathrm{Y}}^{\mathrm{\top }}$ where $\mathrm{\theta }$ is the diagonal matrix of A's singular values $\left({\mathrm{\theta }}_{\mathrm{1}},{\mathrm{\theta }}_{\mathrm{2}}\right.,\left.{\mathrm{\theta }}_{\mathrm{3}},{\mathrm{\theta }}_{\mathrm{4}}\right)\mathrm{\ }$and X and Y are orthonormal bases. $\boldsymbol{\mathrm{K}}\mathrm{=Y\ }\mathrm{\beta }\mathrm{\ }$is one way to express K, where $\mathrm{\beta }\mathrm{=}\left({\mathrm{\beta }}_{\mathrm{1}},{\mathrm{\beta }}_{\mathrm{2}},{\mathrm{\beta }}_{\mathrm{3}},{\mathrm{\beta }}_{\mathrm{4}}\right)$ Thus, reducing(AK)${}^{2}$ also reduces AK:\[{\left(\boldsymbol{\mathrm{AK}}\right)}^{\boldsymbol{\mathrm{2}}}\mathrm{=}{\left(\boldsymbol{\mathrm{AK}}\right)}^{\boldsymbol{\mathrm{T}}}\left(\boldsymbol{\mathrm{AK}}\right)\]
\[\mathrm{=}\left({\mathrm{\beta }}^{\mathrm{T}}{\mathrm{Y}}^{\mathrm{T}}\mathrm{Y}
    \mathrm{\theta }{\mathrm{X}}^{\mathrm{T}}\right)\left(\mathrm{X}\mathrm{\theta
    }{\mathrm{Y}}^{\mathrm{T}}\mathrm{Y}\mathrm{\beta }\right)\]
\[\mathrm{=}{\mathrm{\beta }}^{\mathrm{T}}{\mathrm{\theta }}_{\mathrm{\beta
        }}\]
\begin{eqnarray}
    \mathrm{=}\sum^{\mathrm{4}}_{\mathrm{i=1}}{{\mathrm{\beta }}^{\mathrm{2}}_{\mathrm{1}}{
            \mathrm{\theta }}^{\mathrm{2}}_{\mathrm{1}}}
    \label{Eq11}
\end{eqnarray}
which, when all singular values are set to zero except for the smallest one, is minimal. For instance, if 
${\mathrm{\theta }}_{\mathrm{min}}\mathrm{=}{\mathrm{\theta}}_{\mathrm{4}}$ , then $\mathrm{A}{\mathrm{K}}_{\mathrm{min}}\mathrm{=}{\mathrm{\beta }}_{\mathrm{4}}{\mathrm{\theta }}_{\mathrm{4}}$ ,then 
        $\mathrm{K=}{\mathrm{Y}}_{\mathrm{4}}{\mathrm{\beta }}_{\mathrm{4}}\mathrm{=}\left(\mathrm{x,y,z,1}\right)$ .
Therefore, the coordinates of the triangulated point are:
\begin{eqnarray}
    \mathrm{x}\mathrm{=}{\mathrm{Y}}_{\mathrm{14}}\mathrm{/}{\mathrm{Y}}_{\mathrm{44}},
    \mathrm{y}\mathrm{=}{\mathrm{Y}}_{\mathrm{24}}\mathrm{/}{\mathrm{Y}}_{\mathrm{44}},
    \mathrm{z}\mathrm{=}{\mathrm{Y}}_{\mathrm{34}}\mathrm{/}{\mathrm{Y}}_{\mathrm{44}}
    \label{Eq12}
\end{eqnarray}

Using the confidence $\mathrm{c}$ that gives for each keypoints, we can then represent our equation \ref{Eq10} then we can repeat the process so solves for K.
\begin{eqnarray}
    \mathrm{c\times }\left({\mathrm{A}}^{\mathrm{T}}_{\mathrm{1}}\mathrm{-}\mathrm{X}{
        \mathrm{A}}^{\mathrm{T}}_{\mathrm{3}}\right)\mathrm{K=0,c\times }\left({\mathrm{A}}_{
        \mathrm{2}}\mathrm{-}\mathrm{Y}{\mathrm{A}}^{\mathrm{T}}_{\mathrm{3}}\right)\mathrm{K=0}
    \label{Eq13}
\end{eqnarray}
The result is the coordinates of the triangulated point $\left(\mathrm{x,y,z,c}\right) \mathrm{\ }$like equation \ref{Eq12} and with a factor confidence or accuracy of the model $\mathrm{c}$, where c, ranger from 0 to 1.  Therefore, the threshold value is defined as a parameter for the model to drop coordinates that have a score of $\mathrm{c}$  that is below 0.6 from the data.  In addition, the 2D pose estimation process occasionally incorrectly and with a relatively high confidence detected the occluded key points. In these conditions, the landmark location was incorrectly triangulated and should be removed (i.e excluded markers).  After removing incorrect triangulated data points, the filter process takes care of the missing point through the smooth and gradual transition between the preceding triangulated threshold passband data frames and the existing triangulated threshold stopband data frame. Further details on the filtering process are elucidated in Section \ref{filter}.
Throughout the triangulation process, reprojection errors serve as a qualitative metric for assessing the precision of the 2D image projection onto the 3D world coordinates. The reprojection error is defined as the distance between a pattern keypoint identified in a calibration image and the corresponding world point projected onto the same image. In instances where the overall mean reprojection error exceeds an acceptable threshold, it is strongly advised to exclude the detected keypoints from images with the highest errors and subsequently recalibrate the system\cite{Bradley2010}. For this study, a reprojection error threshold of 8 pixels was set as a criterion for determining the adequacy of the calibration process.
\subsubsection{Data Augmentation}
The data augmentation process leverages established 3D data points derived from the 3D triangulation process to conform to the Biomech-57 Skeleton Marker Set\cite{optitrack23} templates in terms of datatype and marker number. Initially, the 3D triangulated key points, totaling 21 markers as per the OpenPose Body\_25b\cite{Nakano2020} model, are confined by the anatomical landmarks they encompass. However, these key points offer limited insights into the complete kinematics of each body segment's degrees of freedom. Crucial kinematic details, particularly those related to sagittal-plane hip, pelvic, and lumbar movements, cannot reliably be deduced solely from keypoints situated at the shoulders and hips.

Moreover, the inherent limitations of the 3D triangulated key points derived from the OpenPose Body\_25b model encompass various factors, such as the absence of joint centers, variations in anatomical positions due to limb extension or flexion, and the exclusion of hand landmarks. Consequently, the data augmentation process integrates information from the Biomech-57 marker set model\cite{optitrack23}, inferring additional data points from the existing 21 marker key points while modifying their positions. This augmentation is pivotal in overcoming constraints by training two long short-term memory (LSTM) networks\cite{Lindemann2021}. These networks were specifically trained using existing data from subjects performing various actions aided by the Biomech-57 marker set\cite{optitrack23}.

Consequently, these networks have been trained to utilize the 3D positions of the 21 triangulated video key points to predict the 3D positions of the 57 anatomical markers. This expanded set of anatomical markers aligns with those commonly used in marker-based motion capture systems, forming a robust foundation for determining accurate 3D joint kinematics. The adoption of LSTM networks in this context holds promise due to their ability to effectively process time series data. Utilizing LSTM networks is anticipated to significantly enhance the temporal consistency of predicted marker locations, resulting in bolstered accuracy and reliability of the obtained kinematic information.

\subsection{Filter}\label{filter}

Given that most pose estimation algorithms identify key points frame-by-frame, resulting 3D key points' trajectories often exhibit physical unrealism, especially in the presence of misidentified or occluded key points. This poses a significant challenge for biomechanical analysis relying on triangulated 3D key points positions from video data. To overcome these limitations, the study employed a Butterworth filter utilizing the 3D coordinates. Specifically, a fourth-order low-pass Butterworth filter with a cutoff frequency of 6 Hz and zero latency was selected.

By attenuating high-frequency noise while retaining essential movement patterns, Butterworth filters significantly enhance the accuracy and clarity of trajectory data. Notably, the filter proves instrumental in managing sudden, erratic movements within video sequences that are not relevant to the desired analysis. It effectively eliminates noise and outliers in the (x, y, z) data, prioritizing consistent and reliable movement patterns. The frequency selection capability of the Butterworth filter allows for choosing a cutoff frequency, delineating the boundary between preserved and filtered frequencies. In the context of video movement sequences, this feature aids in retaining meaningful movement information while eliminating unwanted high-frequency noise, thereby contributing to the overall accuracy and clarity of trajectory data.

\subsection{Inverse Kinematics and Joint Angle Computation}
The inverse kinematics tool computes a set of joint angles for the model that closely aligns with the recorded experimental kinematics at each time step or frame. To achieve this alignment, OpenSim was used to address an optimization problem using weighted least squares to minimize marker error during motion analysis\cite{Hellsten2021}. To develop this model, the tool employs experimental marker postures to derive the experimental kinematics for the inverse kinematics process.
Marker error refers to the variance between an experimental marker and its virtual counterpart. Each marker holds a specific weight, signifying the degree to which its error should be lessened in the least squares problem. Using these weights, the inverse kinematics computed a vector of generalized coordinates (like joint angles), represented as 'q,' for each frame. This calculation involves determining the weighted sum of marker errors, as reported by Delp et al.\cite{Delp2007}.
\begin{eqnarray}
    \begin{array}{c}
        \mathrm{min} \\
        \mathrm{q}
    \end{array}
    \left[\sum_{\mathrm{i}\mathrm{\in }\mathrm{marker}}{{\mathrm{w}}_{\mathrm{i}}\left
    \|{\mathrm{x}}^{\mathrm{exp}}_{\mathrm{i}}\mathrm{-}{\mathrm{x}}_{\mathrm{i}}\left(
    \mathrm{q}\right)\right\|}\right]\mathrm{\ }
    \label{Eq14}
\end{eqnarray}
Where: $\mathrm{q}$ is a vector of generalized coordinates; ${\mathrm{x}}^{\mathrm{exp}}_{\mathrm{i}}$ is the position of the experimental marker $\mathrm{i}$; ${\mathrm{x}}_{\mathrm{i}}\left(\mathrm{q}\right)$ is the position of the corresponding virtual marker i (which depends on $\mathrm{q}$); ${\mathrm{w}}_{\mathrm{I}}$ is the weight associated with the marker $\mathrm{i}$. Based on the joint position and coordinates information derived from computer vision captures, the output of this process is joint angles
\section{Framework Validation}
The necessity arises to conduct comprehensive validation studies to affirm the efficacy of the proposed computer vision framework for biomechanical analysis. Within the validation section, the framework's accuracy concerning both the motion capture system and its estimation of anthropometric information from each subject needs to be established. The validation process was bifurcated into two segments: assessing the motion capture system and appraising the anthropometric system. The validation studies for the motion capture system will encompass comparisons with established gold standard methods used in biomechanical analysis, such as marker-based motion capture systems. On the other hand, the validation of the anthropometric system primarily focused on evaluating the framework's precision and consistency in capturing and analyzing body measurements. This assessment will involve a comparison of the system's outcomes with ground truth data, involving measurements acquired from a digital scale for body weight and a measuring tape for height, segment length, and circumference.

In validating the framework, ten healthy adult male participants were enrolled in the study. The participants' average age $\mathrm{\pm}$ standard deviation (std), weight $\mathrm{\pm}$ std, and height $\mathrm{\pm}$ std were 27$\mathrm{\pm}$7 years, 77$\mathrm{\pm}$16kg, and 172$\mathrm{\pm}$6cm, respectively. This study received approval from the institutional review board of the University of Michigan Dearborn. Prior to their involvement, participants provided informed consent by signing a consent form. Participants undergo three fundamental tasks in an industrial facility: leaning, bending, and squatting.

Fifty-seven markers were attached to anatomical landmarks on the participants' bodies, as outlined in the Biomech-57 Skeleton Marker Set\cite{optitrack23}. Motion data were captured using the traditional Optitrack motion capture system (OptiTrack Prime, Naturalpoint, Corvallis, OR), which was intended to serve as the ground truth for validating the developed framework in the motion capture system. To evaluate framework performance, a series of tasks were completed by each participant.  One task involved a forward-leaning motion (leaning), simulating the action of reaching for an object. Two forward bending tasks were also completed: a full bend with fully extended knees (bending) and a partial bend with knees flexed (squatting), similar to a self-selected squatting position. Participants performed these tasks while their movements were recorded using both our proposed framework and a traditional motion capture system with reflective markers as mentioned above(OptiTrack Prime, Naturalpoint, Corvallis, OR) motion capture system with markers. This two-pronged approach facilitated a direct comparison of the results obtained from each system for each action.

\subsection{Validation Input}
\begin{figure}[H]
    \centering
    \includegraphics[width=0.8\textwidth]{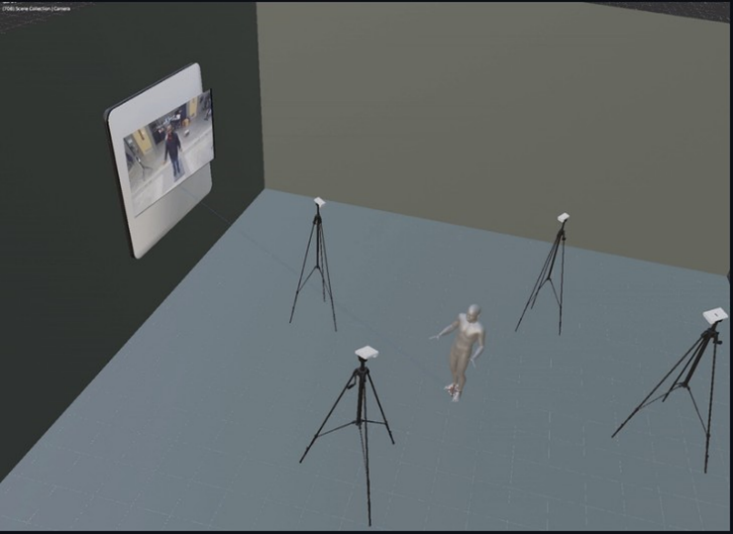}
    \caption{Illustration of the environmental configuration and camera positioning}
    \label{fig:fig11}
\end{figure}

As discussed earlier, the framework relies on video data recorded from each camera during tasks and calibration information specific to each camera. The initial step involves setting up the cameras strategically, as previously mentioned. This positioning is crucial to ensure a comprehensive view of the subject. In this study, four Microsoft Azure Kinect cameras are utilized to capture human motion from diverse angles, as depicted in Figure \ref{fig:fig11}. The Azure Kinect DK, featuring a Time-of-Flight (ToF) depth sensor, enables advanced depth-sensing capabilities within an operational range of 0.25 to 5.0 meters. Complementing this, its high-resolution RGB camera possesses 12 megapixels and a wide field of view, providing a resolution of 512 x 512 pixels. Synchronization of the four cameras ensures complete coverage and seamless coordination of data from all directions. This synchronization enhances spatial awareness and facilitates robust computer vision applications.
The second process involves the extraction of camera calibration information. This pivotal stage in computer vision data collection delineates the coordinate points relative to the 3D worldview. Two fundamental parameters characterize this process: intrinsic camera parameters and extrinsic camera parameters. The acquisition of accurate intrinsic and extrinsic camera parameters is crucial for mapping the camera plane to a real-world plane, thereby enabling a precise estimation of different positions in the image. The execution of camera calibration relies on a checkerboard, a crucial tool that provides calibration points for the cameras. In this study, a printed checkerboard with dimensions of 5x3 vertices and 6x4 squares, where each square measures 95 millimeters (mm), is hosted on an A2 paper. The generous dimensions of the checkerboard contribute to an expansive field of view, covering approximately 9.75 x 9.75 meters. Throughout the calibration process, the checkerboard undergoes various orientations and movements to encompass the entire field of view for each camera. This facilitates the establishment of intrinsic calibration parameters, which include the focal length, optical center, geometric distortion, and skew coefficient of each camera.

The subsequent phase, extrinsic calibration, involves placing the checkerboard within the camera's view, determining rotation, translation of the image plane to the world plane, and optical center parameters (x- and y-axis defining the image plane). Corners are then detected and refined using OpenCV\cite{Culjak2012}. Each camera undergoes calibration through OpenCV's camera calibration algorithm\cite{Zhang2000}. It is imperative to note that the calibration results remain consistent unless the cameras undergo relocation. The robust calibration process ensures accurate and reliable mapping of the image space to the real-world coordinates, essential for subsequent computer vision analyses. The accurate determination of camera parameters, such as focal length, image resolution, and distortion coefficients, is vital. This determination is achieved through a calibration process utilizing calibration patterns, employing the checkerboard placed within the camera's field of view.

\begin{figure}[H]
    \centering
    \includegraphics[width=0.8\textwidth]{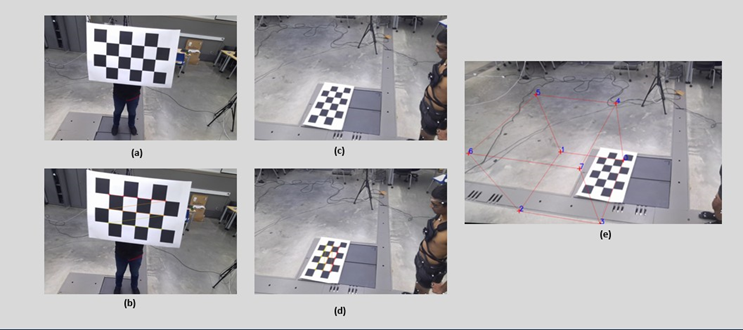}
    \caption{(a). Checkerboard placement for intrinsic calibration. (b). Intrinsic camera calibration result (c). Checkerboard placement for extrinsic calibration (d). Extrinsic camera calibration result (e). A 3D projection of a box based on the calibration result.}
    \label{fig:fig12}
\end{figure}

Strategic placement of the cameras ensures optimal coverage and accuracy. The camera setup for calibration involves two cameras pointed at a 5-degree angle slightly opposite each other (anterior and posterior side) at 3.67 meters apart. The other two are placed on the subject's left and right lateral sides 2.45 meters apart, as shown in Figure \ref{fig:fig11}. Further camera calibration involves placing a known calibration pattern (checkboard) in the camera's view to determine the intrinsic and extrinsic parameters essential for accurate measurement and analysis. The intrinsic parameters include the focal length, pixel scale, and image center, which involves facing the pattern (checkerboard) in the view of each camera (Figure \ref{fig:fig12}a) and taking multiple images from different perspectives. The intrinsic parameters algorithms using OpenCV determine the pattern (Figure \ref{fig:fig12}b) and provide the necessary intrinsic parameters.

On the other hand, the extrinsic parameters determine the position and orientation of each camera relative to the world coordinate system. Therefore, studying the images from multiple cameras reference to the ground, as shown in Figure \ref{fig:fig12}c, allows for calculating the extrinsic parameters using OpenCV to determine the pattern (Figure \ref{fig:fig12}d) and provides the necessary extrinsic parameters. The outcome of the camera calibration process yields DLT and SVD models facilitating the transformation between 2D image coordinates and 3D world coordinates, akin to the procedures detailed in Section \ref{3D_traget}, as depicted in Figure \ref{fig:fig12}e. The duration of the calibration process typically spans 20-25 minutes, contingent upon the quantity of images utilized, which may vary based on the computational capabilities of the computer used for data processing, influencing the speed of the calibration procedure.
\subsection{System Validation}
As previously outlined, the validation of the framework is structured into two distinct components: the assessment of the motion capture system and the evaluation of the volumetric and anthropometric estimation system. The validation process for the motion capture system entails a comparative analysis between established gold standard methods, exemplified by marker-based motion capture systems like the Optitrack system, and the proposed markerless motion capturing system integrated into the framework. The primary aim of motion capture system validation is to confirm the accuracy of marker localization and joint angle estimation. During the validation of the motion capture system, significant attention is directed towards establishing the necessary groundwork to verify marker localization and joint angle estimation accuracy. This verification process initiates with a comprehensive elucidation of the inputs required by the framework, which, as previously discussed, are streamlined to recorded data and camera calibration data. Once the validation input has been established based on data collection procedures specific to each task, the focus shifts to the framework's processing and analysis stages. These stages encompass 2D joint estimation, triangulation, filtering, and inverse kinematic processes. The outcomes obtained, namely marker trajectories and joint angle estimations from the framework, will then be compared with those derived from marker-based motion capture systems using the Optitrack system concurrently for each task. The primary objective of evaluating the volumetric and anthropometric estimation system lies in visually examining the volumetric aspects concerning image sequences and juxtaposing them with ground-truth anthropometric measurements against the framework's estimated outcomes. In the realm of biomechanical analysis, computer vision can plays a critical role in estimating anthropometric parameters such as weight and height, leveraging 3D reconstruction techniques based on body size and shape. Within the framework validation process, a comparison was conducted involving ten male subjects, ground-truth weight recorded from scales with weight estimations generated within the framework, incorporating body segment estimation and measurement. Additionally, it compared the estimated height against measured heights. The weight estimation method integrates various factors, including body density, shape, and vertex parameters. Notably, it accentuates scenarios wherein subjects adopt frontal or T-poses during designated tasks. The process of selecting the pose from the frames is facilitated by anthropometric measurement algorithms. These algorithms systematically scrutinize frames within a video sequence, identifying the frame characterized by the smallest leg segment angle and trunk segment angle. These angles are defined based on research conducted on human postural control by Hwang et al.\cite{Hwang2016}, wherein the angle between the vector hip-to-shoulder and the vector hip-to-leg is considered for frame selection.
\subsubsection{Data Collection and Analysis}
The data collection process varies depending on the specific task at hand, with a focus on capturing video data depicting subjects performing desired activities. These tasks are categorized into three distinct activities: Task 1(Leaning) involves the subject leaning forward, Task 2 (Bending) entails bending to lift a box and placing it back down, and Task 3(Squatting) requires the subject to squat while lifting and dropping boxes. Each task is executed within a timeframe of less than 40 seconds. The video data is meticulously recorded at a frame rate of 30 frames per second, boasting a resolution of 1080p. Following the data collection phase, the subsequent step involves conducting 2D joint estimation on the acquired video data. 2D joint estimation is the technique employed to detect and track key body joints or landmarks within a 2D spatial context. The detection process operates at slightly over 25 frames per second (i.e., 6.25 frames per second for four cameras) on a typical PC equipped with an Nvidia GeForce RTX 2060 graphic card (32 GB memory RAM). Post 2D joint estimation, subsequent stages include triangulation, data augmentation, filtering, and inverse kinematics, contributing to the refinement of accuracy and reliability in CV-based biomechanical analysis studies. As mentioned above, triangulation involves leveraging multiple cameras to precisely track and reconstruct the 3D position of markers on the human body. To optimize the triangulation process and accommodate computational capabilities, a 2D joint estimation confidence threshold of 0.55 and a reprojection error threshold of 10 pixels were selected. Triangulation was completed by utilizing a PC equipped with an Nvidia GeForce RTX 2060 graphics card (32 GB RAM) at approximately 10-15 frames per second. The data augmentation process, which elevates the number of detected markers/key points from 21 to 57, aligns with the Biomech-57 marker set template used in the traditional OptiTrack motion capture system for comparative purposes. This augmentation process becomes imperative to enhance the framework's ability to provide detailed kinematic data and facilitate a meaningful comparison with the marker-based motion capture system. Subsequently, filtering techniques are employed to the motion data, smoothing trajectories and minimizing noise and outliers. The data processing pipeline employs a Butterworth filter to attenuate noise and mitigate fluctuations in the marker/key points data. Following this filtering stage, Inverse kinematics (IK) utilizes filtered motion data to compute joint angles and locations within a biomechanical model, recreating the underlying skeletal movements from surface marker trajectories.
\subsubsection{System Validation Result}
The validation outcome of the Motion Capture System (MCS) comprises 2D joint estimation, the skeletal model, and joint angle plots, as illustrated in Figure \ref{fig:fig14}. In cases where the reprojection error of a 2D joint coordinate exceeds the predefined threshold, the subsequent action involves excluding the corresponding 2D joint. This exclusion varies across frames and is subject to the influence of subject occlusion. Distinctive patterns emerge in various tasks, wherein 2D joint coordinates are computed from excluded data points during the triangulation process may arise where a joint coordinate's reprojection error exceeds the threshold, leading to the exclusion. Across leaning, bending, and squatting tasks, the percentage of excluded markers averages from 4\% to 36\% across all frames, as detailed in Table \ref{tab:table3}. Notably, the mean number of excluded markers is more pronounced in the squatting task, likely attributed to increased object and self-occlusion in tasks 3 and 2 compared to task 1. Although, the data augmentation and filtering process address of the excluded markers, as discussed in Section \ref{3D_target1}. It remains crucial to address occlusion issues, the impact of the reprojection technique, and the effects of filtering on the accuracy of the analysis. The mean reprojection error for bending and leaning hovers around 2.4 pixels, contrasting with an elevation to approximately 5.4 pixels for squatting. Consequently, the triangulation process demands more processing time for the squatting task, primarily due to heightened occlusion and an increased number of excluded markers. These observations underscore the challenges associated with relying solely on the triangulation process, particularly in complex tasks like squatting, where errors and occlusion issues are more prevalent.

\begin{table}[H]
    \centering
    \caption{The triangulation process from OpenPose body\_25b model statistical analysis. The mean absolute reprojection error and the mean number of markers before data augmentation and filtering over time. Mean (mean) and standard deviation (std)}
    \label{tab:table3}
    \begin{tabular}{|c|c|c|c|}
        \hline \multirow{3}{*}{ Tasks } & \begin{tabular}{c} 
        Mean Number of \\
        Excluded \\
        Markers
        \end{tabular} & \multicolumn{2}{|c|}{ Mean Absolute Reprojection Error } \\
        \cline { 2 - 4 } & Mean (\%) & Mean & std \\
        \hline \multirow{2}{*}{ Leaning } & 4.182 & $2.1 \mathrm{px}$ & $1.0 \mathrm{px}$ \\
        & 8.545 & $2.8 \mathrm{px}$ & $1.4 \mathrm{px}$ \\
        \hline \multirow{2}{*}{ Bending } & 4 & $1.9 \mathrm{px}$ & $0.98 \mathrm{px}$ \\
        & 8.545 & $2.8 \mathrm{px}$ & $1.0 \mathrm{px}$ \\
        \hline \multirow{2}{*}{ Squatting } & 22 & $5.4 \mathrm{px}$ & $1.0 \mathrm{px}$ \\
        & 35.636 & $6.1 \mathrm{px}$ & $1.6 \mathrm{px}$ \\
        \hline
    \end{tabular}
\end{table}

\begin{figure}[H]
    \centering
    \includegraphics[width=0.8\textwidth]{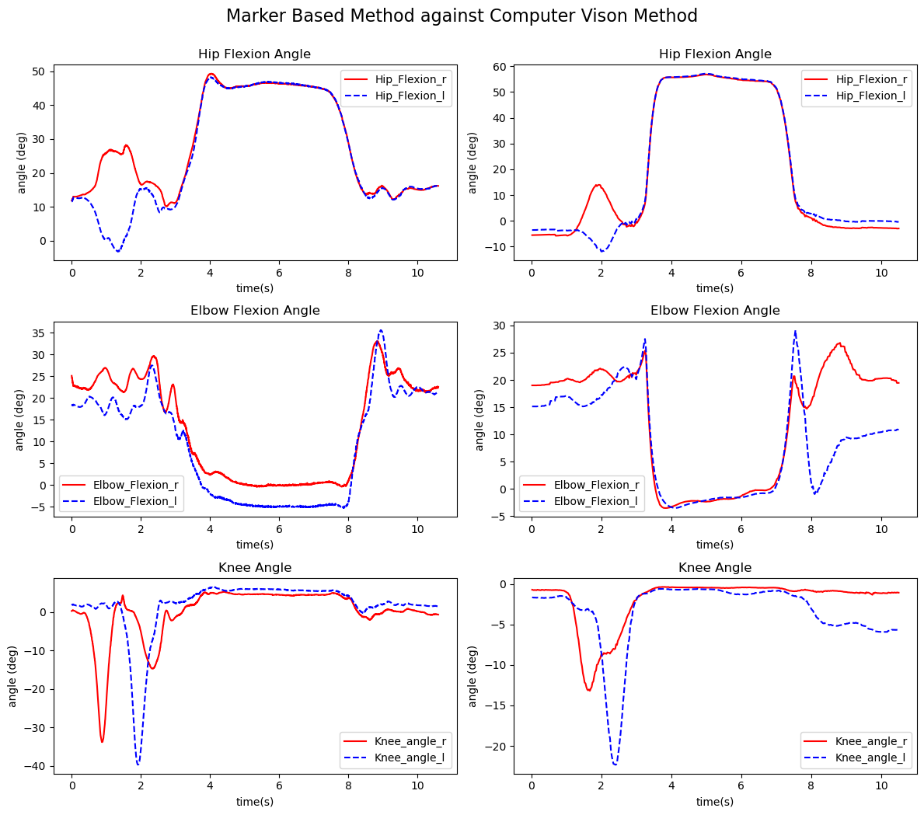}
    \caption{Visual inspection of marker-based method and the framework based on bending task. From top right to bottom right: Is the profile of the hip flexion/extension angle, knee angle, and elbow flexion angle for framework. From top left to bottom left: Is the profile of the hip flexion/extension angle, knee angle, and elbow flexion angle for marker-based method.}
    \label{fig:fig13}
\end{figure}

As illustrated in Figure \ref{fig:fig13}, the joint angle results derived from a bending task using the marker-based method serve as the ground truth against which the framework is compared. Visual analysis reveals a close alignment between the joint angle results obtained from the computer vision method and those derived from the marker-based method, both in terms of the profile of the graph and the magnitude. The hip flexion angle, elbow flexion angle, and knee angle exhibit a similar trend and magnitude in both methods, despite slight differences in exact values and time profiles.

Similar observations are made for the squatting and leaning tasks, aligning with the comparison depicted in Figure \ref{fig:fig13}. The hip flexion angle, elbow flexion angle, and knee angle demonstrate comparable profiles, with a magnitude difference of less than 5 degrees on average between the angles in both the marker-based method and the computer vision method. This consistency across tasks underscores the reliability and accuracy of the present computer vision framework in capturing joint angles, substantiating its effectiveness in biomechanical analysis.

\begin{figure}[H]
    \centering
    \includegraphics[width=0.7\textwidth]{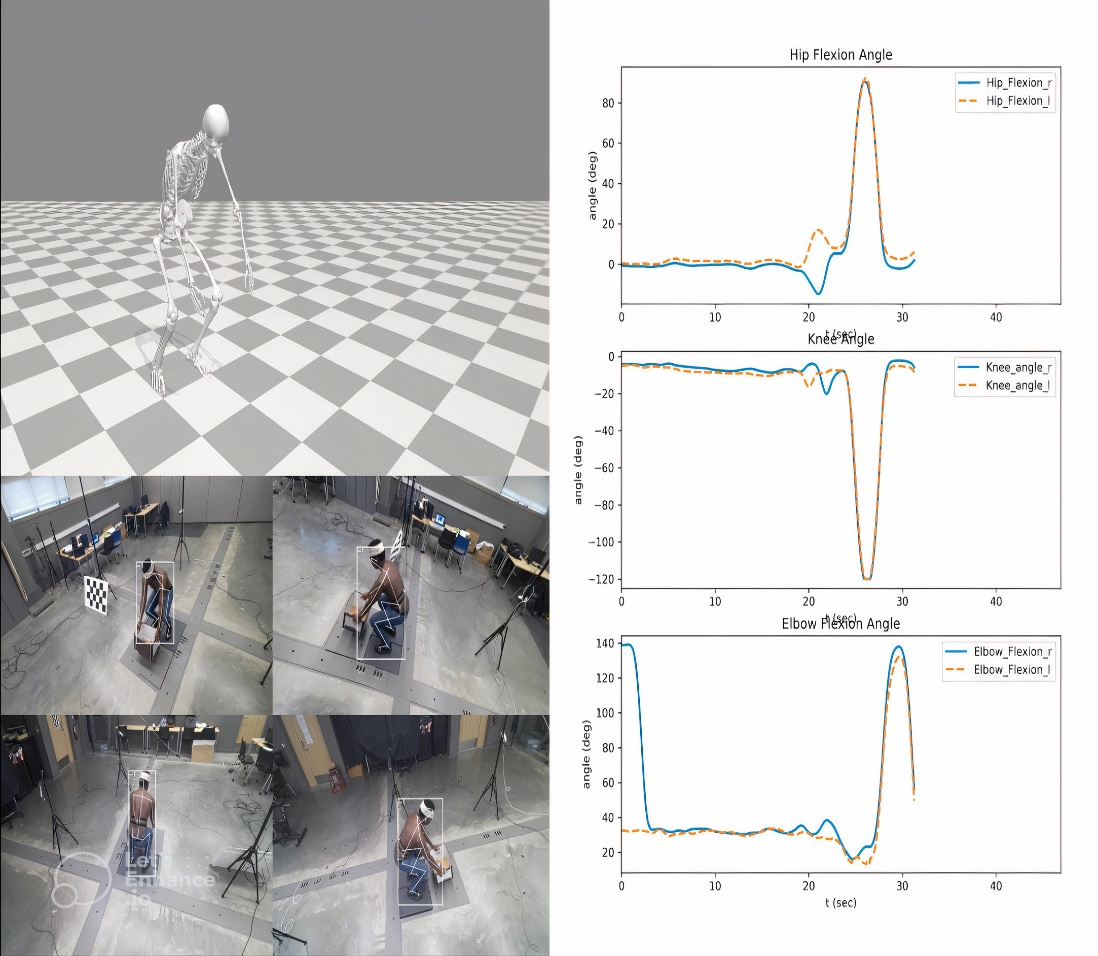}
    \caption{The framework visualization aid for the squatting task at 30 seconds. From top right to bottom right: OpenSim skeleton model in the squatting position and 2D Estimation of the four cameras perspective. From the top left to bottom left: Hip flexion/extension angle, knee angle, and elbow flexion angle.}
    \label{fig:fig14}
\end{figure}

In Figure \ref{fig:fig14}, the concentration of display results from the framework is showcased, specifically focusing on the squatting task. Progressing from the top right to the bottom right, the visual display includes the OpenSim skeleton model in the squatting position, accompanied by the 2D estimations captured from the perspectives of the four cameras. Correspondingly, from the top left to the bottom left, the visual representation encompasses the hip flexion/extension angle, knee angle, and elbow flexion angle. These visualizations offer insights into the accuracy of the framework in capturing joint angles during the squatting task, providing a visual platform to assess the real-time performance of the system.

The validation results revealed an average estimation error of less than 6\% for weight and less than 2\% for height, torso length, arm length, and shoulder-to-shoulder length across the ten subjects in the this study. However, the weight estimation error appeared relatively large, primarily attributable to substantial variations in estimating body density among the subjects. Despite this discrepancy, these findings underscore a commendable level of accuracy in the framework's capacity to estimate height, relative segment measurements, and weight. This indicates a high level of accuracy in the framework's ability to estimate height, relative segment measurement, and weight, showcasing its potential as a reliable tool for anthropometric measurements in the context of biomechanical analysis.

\begin{figure}[H]
    \centering
    \includegraphics[width=0.7\textwidth]{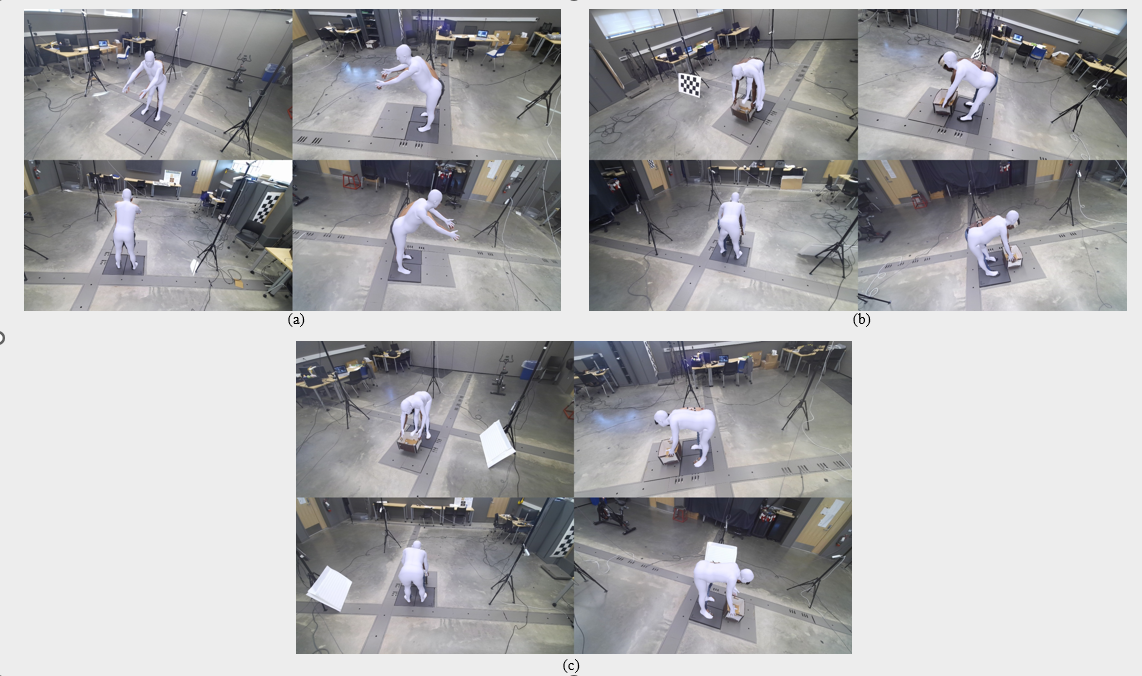}
    \caption{(a) Volumetric estimation overlaid on the four camera views depicting the leaning task. (b) Volumetric estimation overlay on the four camera views illustrating the bending task with mesh deformation. (c) Volumetric estimation overlay on the four camera views showcasing the bending task with improved estimation}
    \label{fig:fig15}
\end{figure}
Figure \ref{fig:fig15} displays the volumetric estimation outcomes superimposed on the subject's body shape, as obtained through the framework. In Figure \ref{fig:fig15}(a), the colored overlay depicts the estimated body volume while the subject engages in a leaning task. Figure \ref{fig:fig15}(b) illustrates a similar representation, but for a distinct task involving bending to lift a box. However, the volumetric mesh experiences considerable deformation during the bending task due to trunk flexion, limb movement, and occlusion by the box, consequently impacting estimation accuracy. Subsequently, in Figure \ref{fig:fig15}(c), enhancements were made to the results depicted in Figure \ref{fig:fig15}(b) by augmenting similar occluded data to the training dataset and retraining the framework. This iterative process aimed to refine the accuracy of body volume estimation, particularly during bending tasks.
\begin{figure}[H]
    \centering
    \includegraphics[width=0.8\textwidth]{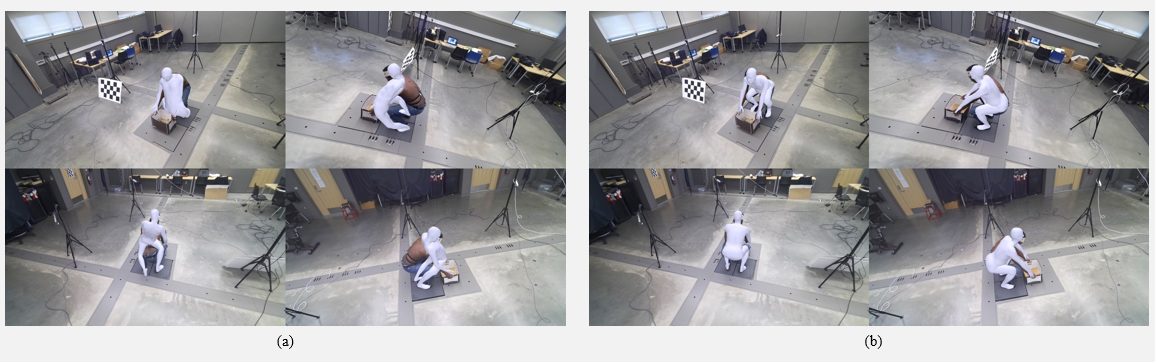}
    \caption{(a) Volumetric estimation overlaid on the four camera views for the squatting task with mesh deformation. (b) Volumetric estimation overlay on the four camera views for the squatting task with enhanced estimation.}
    \label{fig:fig16}
\end{figure}
In Figure \ref{fig:fig16}(a), the estimated body volume overlay is shown while the subject performs a squatting task. Figure \ref{fig:fig16}(a) mirrors Figure \ref{fig:fig15}(b), depicting volumetric mesh deformation resulting from trunk, limb, and knee flexion, as well as occlusion by the box, thereby affecting estimation accuracy. In  Figure \ref{fig:fig16}(b), improvements are demonstrated over the results presented in  Figure \ref{fig:fig16}(a). This was achieved by utilizing the refined model and enhancing detection resolution for landmarks or keypoints, thereby refining the accuracy of volumetric estimation.
\begin{table}[H]
    \centering
    \caption{Framework results in this study of a walking task in comparison to Pose2Sim, Theia3D, and Xsens focus on mean error (ME), and root-mean-square error (RMSE)}
    \label{tab:table4}
    \begin{tabular}{|c|c|c|c|}
    \hline 
    \textbf{Joint} & \textbf{Method} & \multicolumn{2}{|c|}{\textbf{Flexion/Extension}} \\
    & & \textbf{ME} ($\left.{}^{\circ}\right)$ & \textbf{RMSE} $\left(^{\circ}\right)$ \\
    \hline 
    \multirow{4}{*}{\rotatebox{45}{Knee}} & Pose2sim\cite{Pagnon2021} & 1.92 & 5.1 \\
    & Theia3D\cite{Kanko2021} & - & 3.3 \\
    & Xsens\cite{Zhang2013} & 1.87 & - \\
    & This study & 1.74 & 4.2 \\
    \hline  
    \multirow{4}{*}{\rotatebox{45}{Hip}} & Pose2sim\cite{Pagnon2021} & 3.04 & 5.6 \\
    & Theia3D\cite{Kanko2021} & - & 11 \\
    & Xsens\cite{Zhang2013} & 2.47 & - \\
    & This study & 1.92 & 4.8 \\
    \hline  
    \multirow{4}{*}{\rotatebox{45}{Elbow}} & Pose2sim\cite{Pagnon2021} & 3 & - \\
    & Theia3D\cite{Kanko2021} & - & - \\
    & Xsens\cite{Zhang2013} & - & - \\
    & This study & 2.1 & 6.3 \\
    \hline
    \end{tabular}
\end{table}

\section{Discussion}
\subsection{Findings and Analysis}
In this investigation, a computer vision-based biomechanical framework was devised to facilitate 2D and 3D body segment tracking and anthropometry extraction. The current iteration of the framework demonstrates proficiency in estimating a two-dimensional pose, volumetric shape, and geometry of human actions, as well as providing crucial metrics such as total body weight, height, and the weights and relative lengths of distinct body segments (arms, head, torso, thighs, legs). This framework serves as an automatic and non-invasive tool for identifying human physical characteristics, thereby contributing to the prompt assessment of an individual's instantaneous physical status. The developed computer vision framework eliminates the need for physical markers or tracking devices, offering essential anthropometric data requisite for comprehensive human motion analysis.

The analyses encompassed a dataset featuring ten subjects, subject to diverse statistical assessments aimed at validating the framework's efficacy. Various tasks, including walking, leaning, bending, and squatting, were incorporated to evaluate the framework's performance, drawing comparisons with a parallel marker-based motion capture method. Key metrics analyzed involved mean absolute error, mean error, standard deviation, standard deviation ratio, and percent error.

In the assessment of joint angles, mean absolute angle errors for hip flexion, elbow flexion, and knee angles ranged between 1$\mathrm{{}^\circ}$ and 7$\mathrm{{}^\circ}$, 2$\mathrm{{ ^\circ}}$ and 6$\mathrm{{}^\circ}$, and 0$\mathrm{{}^\circ}$ and 3$\mathrm{{}^\circ}$, respectively. The outcomes indicate a more pronounced mean absolute angle error in degrees for challenging tasks, such as squatting, in contrast to less demanding activities like bending due to a more pronounced movement in the trunk flexion, limb movement, and knee flexion. The plot in Figure \ref{fig:fig13} illustrates slight differences in exact values and time profiles; however, the hip flexion angle, elbow flexion angle, and knee angle exhibit a similar trend and magnitude in both the marker-based method and our framework. Moreover, the framework was benchmarked against existing markerless and marker-based methods for joint angle analysis during walking tasks. The framework's mean error for hip, knee, and elbow angles ranged from 1.7 to 2.5 degrees, signifying a high level of accuracy in capturing and analyzing  joint angles. In contrast, the Pose2sim\cite{Pagnon2021}, Theia3D\cite{Kanko2021}, and Xsens\cite{Zhang2013} consistently underestimate the joint angles by a range of 1.8 to 3 degrees as depicted in Table \ref{tab:table4}. Framework results in this study of a walking task in comparison to Pose2Sim, Theia3D,and Xsens focus on mean error(ME), and root-mean-square error(RMSE).

The anthropometric measurements demonstrate the precision of the framework, with the height and segment length measurements exhibiting greater accuracy than weight estimation. This observation is attributed to the framework's enhanced capability to extract information pertaining to height and length compared to weight. Notably, weight estimation is influenced by the product of volume and human density, which varies among individuals. Despite the improvement achieved by incorporating the Body Mass Index (BMI) constraint, providing a more accurate estimate compared to using the average density of $985\frac{kg}{m^3}$, there remains an opportunity for refinement. This can be achieved by incorporating a statically model-based\cite{Bartol2022} height estimation and incorporating weight variation to predict human density, contingent upon sufficient data. The findings show that the mean absolute error (MAE) of estimation for height and weight is less than 2\% and less than 6\%, respectively. This discrepancy highlights how much more accurate the framework is at estimating height than weight. These results highlight the framework's potential as a reliable instrument for anthropometric measurements in the context of biomechanical analysis, which has consequences for the framework's dependability in practical settings.

\subsection{Framework Potential Applications}
The framework methods in biomechanics allow for non-invasive, precise, and often real-time analysis of human movement, enabling advancements across various fields, from healthcare to sports, and from design to rehabilitation\cite{Colyer2018}.

One area of application of this computer vision framework to biomechanical analysis is in gait analysis. The computer vision method is used to capture the gait pattern information using developed algorithms. Understanding gait analysis is vital in the clinical assessment of patients with disease conditions like Parkinson's disease and in athlete injury prevention through an understanding of gait patterns in sports biomechanics.

Another area of application of this study is motion capture. This is useful in body movement tracking of athletes in sports like basketball and soccer, and the tracked motion capture information can be used to improve the athlete's technique in the sport and reduce the risks of injury during sports performance. Also, in rehabilitation, patients are monitored and guided through exercise by analyzing their movement information captured through motion capture algorithms.

Finally, another area of application is in workplace safety and ergonomic product designs. Using computer vision to capture the motion and posture information of the workers in the workplace will enable workplace assessment and design and will prevent ergonomic injuries in the workplace.

\subsection{Limitations and Future Research and Development}

The capture design employed in the study imposed limitations on the field of view, confining observations to a single subject. The current framework centers its analysis on the movement of an isolated individual, such as an injured worker. Expanding the scope of analysis to encompass multiple athletes engaged in simultaneous activities, particularly in team sports, combat sports, and races, could yield valuable insights. Addressing this limitation entails broadening the camera setup to cover a wider field of view and integrating advanced tracking algorithms capable of concurrently managing multiple subjects. This methodology involves two steps: initially sorting the detected individuals during tracking by triangulating those with a reprojection error below a defined threshold, as opposed to exclusively selecting the one with the least error; subsequently, tracking the triangulated individuals over time.

In addition, several challenges stem from occlusion problems of varying degrees depending on the tasks. The primary occlusion issues encountered involve self-occlusion and object-to-background occlusion. Self-occlusion occurs when body parts or joints are obstructed by other body parts, such as when an arm covers the chest during a reaching movement. Object-to-background occlusions occur when environmental elements obstruct objects or body parts, as seen in industrial settings with walls, furniture, or boxes. Adjusting camera positions lower and closer to the ankles using marker-based methods could have partially mitigated this issue. However, due to OpenPose's reliance on optimal visibility of the entire body for effective operation, this was not feasible in this instance.

The complexity of biomechanical analysis and the limitations of existing technology prevent the framework from providing wholly accurate joint coordinates relative to anatomically precise joint positions. Key components of significance are the OpenSim models and the 2D pose estimation. While the OpenPose Body\_25b identifies only 25 key points representing anatomical joint positions, a minimum of 47 joints exists. To address this, a recurrent neural network (LSTM) was utilized to generate 3D positional markers for anatomical landmarks. However, discrepancies arise in OpenPose key points' locations concerning joint centers, particularly when body parts are fully extended or flexed. Additionally, hand landmarks are absent in the OpenPose Body\_25b, presenting challenges in estimation without sufficient data.

The majority of the study's tasks were centered around sagittal plane symmetry. It is necessary to assess how well the model performs in different rotation planes and in more difficult tasks. Although the framework adeptly manages tasks such as leaning forward, bending, and squatting with minimal issues, its efficacy in intricate movements like jumping or twisting, which involve significant off-plane motion, requires careful examination. Quick motions and occlusions significantly impact joint tracking accuracy. Therefore, comprehensive assessments of tasks involving diverse 3D movements, challenging 2D pose estimation scenarios (e.g., upside-down individuals during flipping exercises), and unique environmental factors (e.g., swimming with distinct refractive indexes and limb extremity splashes) are essential. Additionally, evaluating the model's performance in disciplines with distinct outfits and substantial occlusions, such as motorbiking or fencing, would yield valuable insights.

Another potential weakness lies in the limited size of the subject dataset, comprising only ten male subjects. Expanding the dataset to include diverse demographics and a larger sample size could enhance the generalizability of the framework's performance across a broader population.The framework's clinical efforts have been instrumental in establishing the validity and accuracy of joint angle analysis. Nevertheless, in biomechanical analysis, considering joint moments, forces, and overall biomechanical movement dynamics is crucial. This necessitates factoring in elements such as muscle activation patterns, joint kinetics, and the transfer of kinetic energy. Further expanding the framework's capabilities to account for joint moments, forces, and loads is imperative in advancing its biomechanical analysis scope.

Future experiments could explore refining the framework's performance by expanding the dataset to include a more diverse group of subjects, considering gender, age, and physical attributes. Additionally, further investigation into the nuances of joint angle variations during challenging tasks could provide insights for optimizing the framework's robustness in various biomechanical scenarios.

\section{Conclusion}

The computer vision framework designed for biomechanical analysis demonstrates significant potential in assessing and analyzing human movements and postures. Rigorous validation processes are implemented to ensure the accuracy and reliability of this framework. The validation involves a comprehensive comparison with gold-standard marker-based motion capture systems and ground truth measurements. Specifically, the accuracy and reliability of joint position estimation are validated by comparing results with joint angles obtained from marker-based systems. The joint angles derived from the markerless approach are scrutinized against those from marker-based systems to establish a high level of agreement. Consistency in joint angle profiles for the hip, knee, and elbow indicates that the computer vision framework accurately and reliably estimates joint positions and angles.

In addition to joint position validation, the accuracy and reliability of the anthropometric system are assessed by comparing body measurements from the computer vision framework with ground truth measurements acquired using a digital scale and measuring tape. Comparative analyses with existing markerless motion capture systems further affirm the framework's accuracy and reliability in pose estimation and body measurement analysis. Results indicate superior accuracy in joint position estimation compared to some existing software.

While the framework demonstrated reliability across various tested conditions, speed considerations were not explicitly addressed. Timely analysis of individual movement activities in an industrial environment appears feasible, although achieving real-time analysis remains challenging at the current stage of framework development. Future development and refinement of the computer vision framework are anticipated to enhance its accuracy, reliability, and speed.

\end{document}